\documentclass[journal,twoside,web]{ieeecolor}
\usepackage{generic}
\usepackage{amsmath,amssymb,amsfonts}
\usepackage{algorithmic}
\usepackage{algorithm}
\usepackage{array}
\usepackage[caption=false,font=normalsize,labelfont=sf,textfont=sf]{subfig}
\usepackage{textcomp}
\usepackage{stfloats}
\usepackage{url}
\usepackage{verbatim}
\usepackage{graphicx}
\usepackage{cite}
\usepackage[backref]{hyperref}
\usepackage{makecell}

\def\BibTeX{{\rm B\kern-.05em{\sc i\kern-.025em b}\kern-.08em
    T\kern-.1667em\lower.7ex\hbox{E}\kern-.125emX}}
\begin{document}

\title{Segmentation of Parotid Gland Tumors Using Multimodal MRI and Contrastive Learning}

\author{Zi'an Xu, Yin Dai, Fayu Liu, Boyuan Wu, Weibing Chen, Lifu Shi
\thanks{This research was funded in part by the National Natural Science Foundation of China grant number 61902058, in part by the Central University Basic Research Fund of China grant number N2019002, in part by the Natural Science Foundation of Liaoning Province grant number No. 2019-ZD-0751, in part by the Central University Basic Research Fund of China grant number No. JC2019025, in part by the Medical Imaging Intelligence Research grant number N2124006-3, in part by the National Natural Science Foundation of China grant number 61872075.}}

\maketitle

\begin{abstract}
Parotid gland tumor is a common type of head and neck tumor. Segmentation of the parotid glands and tumors by MR images is important for the treatment of parotid gland tumors. However, segmentation of the parotid glands is particularly challenging due to their variable shape and low contrast with surrounding structures. Recently deep learning has developed rapidly, which can handle complex problems. However, most of the current deep learning methods for processing medical images are still based on supervised learning. Compared with natural images, medical images are difficult to acquire and costly to label. Contrastive learning, as an unsupervised learning method, can more effectively utilize unlabeled medical images. In this paper, we used a Transformer-based contrastive learning method and innovatively trained the contrastive learning network with transfer learning. Then, the output model was transferred to the downstream parotid segmentation task, which improved the performance of the parotid segmentation model on the test set. The improved DSC was 89.60\%, MPA was 99.36\%, MIoU was 85.11\%, and HD was 2.98. All four metrics showed significant improvement compared to the results of using a supervised learning model as a pre-trained model for the parotid segmentation network. In addition, we found that the improvement of the segmentation network by the contrastive learning model was mainly in the encoder part, so this paper also tried to build a contrastive learning network for the decoder part and discussed the problems encountered in the process of building.
\end{abstract}

\begin{IEEEkeywords}
contrastive learning, transfer learning, parotid gland tumor, image segmentation, decoder.
\end{IEEEkeywords}

\section{INTRODUCTION}
\IEEEPARstart{G}{lobally}, there are approximately 400,000 to 600,000 new cases of head and neck cancer each year, resulting in 223,000 to 300,000 deaths \cite{s1}. Salivary gland tumors constitute 3\% to 12\% of head and neck tumors \cite{s2}. About 80\% of salivary gland tumors originate from the parotid gland, which is the largest of the three pairs of salivary glands in the human body \cite{s3}. Parotid gland tumors (PGTs) are therefore the most common type of salivary gland tumors \cite{s4, s5}. Surgery is one of the main treatments for PGTs, and the choice of surgical approach is closely related to the location of the tumor \cite{s3}. Radiotherapy also requires accurate segmentation of tumors and organs at risk (OARs) to be able to plan and deliver a sufficient dose to the tumors while minimizing side effects to the OARs \cite{s6}. Therefore, the segmentation of parotid glands and tumors is important for the treatment of PGTs.

Through MRI, patient’s anatomy can be evaluated without the use of ionizing radiation \cite{s7}, and more information can be provided about the internal structure, localization, and relationship to other tissues of PGTs than other imaging methods \cite{s8}. MRI therefore plays a pivotal role in the noninvasive evaluation of PGTs, and any method that has the potential to improve the accuracy and efficiency of MRI-based preoperative diagnosis would greatly benefit patients with PGTs \cite{s9}. However, manual segmentation of the parotid regions and tumors by MR images is a subjective, time-consuming, and error-prone process \cite{s1}. Automated image analysis methods allow for consistent, objective, and rapid diagnosis of PGTs \cite{s10}, and significantly reduce the time clinicians must spend on this task \cite{s6}. However, segmentation of the parotid glands is particularly challenging due to their variable shape and low contrast with surrounding structures \cite{s11}.

In recent years, deep learning has developed rapidly in the computer vision (CV) field. Especially, after the success of Transformer \cite{1} in the CV field, a lot of excellent work has been published \cite{2, 3, 4, 5, 6}. Compared with traditional digital image processing methods, deep learning has great advantages in processing complex image problems \cite{7}, which makes it also has many important applications in medical image processing. There are already many deep learning methods that can handle the segmentation of medical images \cite{8, 9, 10, 11, 12}. However, most of the current medical image segmentation methods using deep learning are still based on supervised learning methods, which means that training these models requires a large amount of labeled medical image data.

Compared with natural images, on the one hand, medical images are scarce \cite{13}, which is due to the high cost of capturing medical images, the complexity of the capture process, and the small number of samples. Therefore, for network training, every medical image is valuable. On the other hand, labeling medical images is costly \cite{14}, since medical images require specialized clinicians for labeling and their content is more complex and requires more time and effort, especially for the pixel-level labeling required for segmentation tasks. Although recently there are some methods to simplify the labels needed for segmentation, for example, in 2022 Mingeon Ju et al. suggested that a small number of points could be used instead of the full segmentation labels \cite{15}. But labeling is still a time-consuming and laborious task that requires specialized clinicians. In summary, the situation is that medical images are inherently valuable and the cost of labeling is high, which requires algorithms that can make full use of those medical images even if they are not labeled. Compared with supervised learning methods, unsupervised learning methods that do not require labels are better at solving this problem.

Contrastive Learning, as an unsupervised learning method, has evolved rapidly in recent years. In 2018, Aaron van den Oord et al. proposed CPC \cite{16}, using a generative pretext task that uses the information in front of a sequence to predict the information behind it. It is suitable for processing time series. In the same year, Zhirong Wu et al. proposed InstDisc \cite{17}, using an instance discrimination task that has been a great inspiration for subsequent work on contrastive learning. In 2019, Mang Ye et al. proposed InvaSpread \cite{18}, which pioneered the use of an end-to-end learning method to train the instance discrimination task. In the same year, Yonglong Tian et al. proposed CMC \cite{19}, which used multimodal data for contrastive learning, reflecting the flexibility of contrastive learning. Moreover, MoCo v1 \cite{20} proposed by Kaiming He et al. in the same year was the first time that contrastive learning achieved accuracy comparable to that of supervised learning in several mainstream tasks in the CV field. In 2020, BYOL \cite{21}, proposed by Jean-Bastien Grill et al. broke the usual contrastive learning paradigm by making contrastive learning free from model collapse even without the use of negative samples.

In summary, as a promising unsupervised learning method, contrastive learning is well suited for use in the field of medical image processing. The global prevalence of COVID-19 in recent years has produced many CT and MR images of the lung. But the disease epidemic has also put a lot of pressure on health care workers, making most of these images unlabeled. Therefore, several works have been started to try to solve the medical image classification problem of COVID-19 using contrastive learning methods \cite{22, 23, 24, 25}. Besides, Bin Li et al. also used contrastive learning to solve the classification problem of high-resolution whole slide images \cite{26}. These works have proven the feasibility of contrastive learning for processing medical images. In this paper, we take up the work of our previous paper \cite{27} and use contrastive learning to process unlabeled images in the parotid dataset. And we innovatively use transfer learning to train contrastive learning networks. Compared with the pre-trained model with supervised learning, the obtained contrastive learning model is more suitable for transferring to the downstream parotid segmentation task, which further improves the accuracy of the parotid segmentation model.

The rest of this paper is organized as follows. Section II describes our previous work and the contrastive learning methods related to this paper to help the readers understand the content of the subsequent paper. Section III describes the specific methods and datasets used in this paper. Section IV presents the relevant experiments and experimental results, followed by our discussion. Finally, our work is summarized in Section V.

\section{RELATED WORK}
\subsection{Swin-Unet using supervised learning model for transfer learning}
This study is based on our previous research results, so a brief introduction is needed. In the previous paper \cite{27}, we collected multicenter multimodal MR images of the parotid gland. Swin-Unet \cite{28} based on Transformer \cite{1} was used. The MR images of STIR, T1, and T2 modalities were combined into three-channel images to train the network. Segmentation of the parotid and tumor region of interest (ROI) was achieved. The model has a DSC of 88.63\%, MPA of 99.31\%, MIoU of 83.99\%, and HD of 3.04 in the test set. Then we designed the following four experiments to further verify the network performance.

First, we trained eight other classical deep learning segmentation networks by using the same data, including U-Net \cite{29}, UNet++ \cite{30}, MA-Net \cite{31}, LinkNet \cite{32}, PSPNet \cite{33}, PAN \cite{34}, DeepLabV3 \cite{35}, and DeepLabV3+ \cite{36}. Eventually, it was found that Swin-Unet performed the best. This is likely since the backbone network of Swin-Unet is Swin Transformer \cite{37}. Compared with other segmentation networks using convolutional neural networks as the backbone network, the Transformer structure has a stronger global modeling capability.

Second, in order to verify whether training the network using multimodal data would bring better results than using only single modal data, a comparison experiment using multimodal data and a single modal data was designed. The three unimodal groups were trained with three copies of one of the three modalities, STIR, T1, and T2, as the three-channel images input the network. The multimodal group used all three modalities simultaneously as the three-channel images input the network. The results of the experiments showed that using multimodal data to train the network does lead to better results.

Third, the pre-trained model used for transfer learning is trained with natural images in ImageNet with R, G, and B as the three channels. The images of the parotid gland used for the experiments have three channels: STIR, T1, and T2. Although they are all three-channel images, they have significant differences in physical meaning and data distribution. To verify whether transfer learning can still be useful, a comparison experiment with and without transfer learning was designed. The results of the experiments show that transfer learning can still be effective. This may be because transfer learning can compensate for the lack of inductive bias in the Transformer structure.

Fourth, we designed experiments to validate the training method with multicenter data. The results of the experiments show that the results are better when the train and test sets originate from the same center, and the accuracy is comparable to mixing the multicenter images first and then randomly selecting the train and test sets from them. The results are worse when the train and test sets originate from different centers. Therefore, it is concluded that data from different centers vary greatly in deep learning. The training set of the network should contain data from multiple centers so that the trained model can maintain high accuracy on data from different centers.

Based on the conclusions drawn from the above research results, the experiments in this paper will still use Swin-Unet to implement the segmentation of parotid gland and tumor. The multimodal data are still used to form three channels, and the multicenter data is mixed and then randomly selected for the train and test sets. However, for the following two reasons, the pre-trained model for transfer learning of Swin-Unet in the experiments of this paper will be trained using the contrastive learning method.

First, as mentioned in Section I, medical image data is scarce and precious, while labeling is a time-consuming and laborious task. To save time and effort, the parotid data set used in this experiment was labeled only on the side with the tumor. Therefore, the image data on the side without tumor was not used in the previous study. Contrastive learning, as an unsupervised learning method, does not require manual labeling. Therefore, the contrastive learning method can be used for almost all images in the parotid data set.

Second, the pre-trained model used for transfer learning in the experiments of the previous paper was obtained from the Swin Transformer network with supervised learning on the ImageNet dataset. Although the experiments show that better results are achieved using transfer learning, since the R, G, and B channels of the natural images are significantly different from the STIR, T1, and T2 channels of the MR images in this experiment, we believe that the better results are largely due to the pre-trained network compensating for the lack of inductive bias in the Transformer. Therefore, we conjecture that better results may be achieved if pre-trained models are obtained directly using contrastive learning on the parotid dataset and then transferred to the downstream parotid segmentation task.

\subsection{InstDisc and InvaSpread}
Contrastive learning is an unsupervised learning method. The learning goal of a classical contrastive learning network is to make positive sample pairs as close as possible and negative sample pairs as far away as possible in the feature space. And among them, positive and negative sample pairs need to be defined artificially, which is the role of the pretext task in contrastive learning. The definition of positive and negative sample pairs can be given by setting up some logically simple tasks that have no practical use. This will enable the network to learn from the data even in the absence of labels, thereby showing better performance in the transfer of downstream tasks.

Zhirong Wu et al. proposed InstDisc in 2018 \cite{17}. This method proposed a pretext task called instance discrimination. Specifically, the instance discrimination task takes an image and does different data enhancements to get two images that look different. However, the semantics of these two images should be the same as the original image, so these two images form a positive sample pair. The task considers that all images except this one are not in the same class as this one, so this image and the other images form several negative sample pairs.

Another major contribution of InstDisc is the proposed use of a memory bank to store negative sample image features. Specifically, all images are first encoded to obtain 128-dimensional feature vectors, and then these feature vectors are stored in the memory bank. Several features are randomly selected from the memory bank as negative samples for the next training. After calculating the loss, the encoder is updated using gradient backpropagation. The old features in the memory bank are replaced by the new features calculated with the updated encoder. This enables feature and encoder updates.

Of course, using a memory bank is not the only way to train for the instance discrimination task. An end-to-end learning approach is used in the InvaSpread \cite{18} method proposed by Mang Ye et al. in 2019. This approach randomly takes a mini-batch of images from all the dataset images and then goes through the encoder to get the image features as positive and negative sample pairs. After calculating the loss, the encoder is updated directly using gradient backpropagation. In the next learning round, another mini-batch of images is randomly taken out, and the new image features are obtained by the updated encoder. This also enables feature and encoder updates.

In summary, the instance discrimination task is a logically simple and efficient enough task to be used extensively in later studies. The two papers, InstDisc and InvaSpread, also set the stage for very much subsequent contrastive learning work.

\subsection{MoCo v1 and MoCo v2}

MoCo v1 \cite{20} is a contrastive learning architecture proposed by Kaiming He et al. in 2019, which also uses the instance discrimination task. Compared with InstDisc and InvaSpread, MoCo v1 innovatively uses queues to store negative sample features and uses momentum encoders, thus achieving better results.

The authors of MoCo v1 argue that the contrastive learning approach using the instance discrimination pretext task can be viewed as a dictionary look-up task. Two different images are obtained after data enhancement. The features obtained from one of the two images after encoding can be considered as a query, and the other one is the key corresponding to this query. This key together with other keys of negative samples form a dictionary. Then contrastive learning is to train the network to find the key that is closest to the query.

The dictionary look-up task requires a large and feature-consistent dictionary for the network to have good training results. However, InstDisc and InvaSpread are either limited by the size of the dictionary or the poor feature consistency of the dictionary. Specifically, InvaSpread is trained end-to-end, and the size of the constructed dictionary is equal to the size of the mini-batch. But due to the limitation of GPU hardware memory, it is not possible to set the mini-batch to a large size. Therefore, this end-to-end learning approach is limited by the size of the dictionary. InstDisc uses a memory bank to store image features and randomly picks features from the memory bank during training, so the size of the dictionary is no longer limited by the batch size. But this approach will cause the network to train a full epoch to update all the features in the memory bank. And the features randomly sampled at one moment in the next epoch are likely to come from the encoder at a different moment in the previous epoch. Therefore, the consistency of features is poor.

MoCo v1 mainly proposes two innovations to solve the problems of small dictionaries and poor feature consistency, respectively. Firstly, MoCo v1 uses a queue to store image features so that the dictionary size is not limited by the batch size. Another advantage of using a queue is that the new features obtained by the encoder at a certain moment are queued in, and the features that are queued out are the ones that are the most inconsistent produced by the encoder with the longest interval from that moment. Secondly, the key encoder of MoCo v1 uses momentum update. The parameter ${k_t}$ of the key encoder at a certain moment is obtained by weighting the sum of the parameter ${k_{t - 1}}$ of the previous moment and the parameter ${q_t}$ of the query encoder at the current moment, as shown in (1). Where $m$ is the momentum coefficient, the larger the momentum coefficient, the slower the key encoder is updated. By using a very large momentum coefficient, the key encoder is updated slowly, thus ensuring the consistency of the features generated by the encoder.

\begin{equation}
\label{equ_1}
{k_t} = m \cdot {k_{t - 1}} + (1 - m) \cdot {q_k}
\end{equation}

In summary, MoCo v1 constructs a large and consistent dictionary, thus dramatically improving the model performance of contrastive learning. MoCo v1 is a landmark work because it meets or exceeds supervised learning in several downstream tasks, demonstrating the feasibility of unsupervised learning.

SimCLR was proposed by Ting Chen et al. in 2019 \cite{38}. They found that adding a projector consisting of linear layers and ReLU activation functions after the encoder resulted in a significant improvement in the performance of the model. Besides, a larger batch size and more data enhancement will also improve the performance of the model. These tricks in SimCLR are universally applicable. Adding these tricks to MoCo v1 yields MoCo v2 \cite{39}.

One of the great advantages of MoCo v1 and MoCo v2 over other contrastive learning methods is the lower hardware requirements. SimCLR uses an end-to-end learning approach and sets the maximum batch size to 8192, which is not enough memory for a typical GPU to put down such a large batch size. MoCo v2 uses only a batch size of 256 to get better results than SimCLR with a batch size of 4096. 

\subsection{BYOL and MoCo v3}
BYOL was proposed by Jean-Bastien Grill et al. in 2020 \cite{21}. The method breaks the usual paradigm of contrastive learning and makes contrastive learning less dependent on negative samples. In the previous contrastive learning method, if only positive samples are used, the encoder of the network only needs to make the features of the positive samples close to each other. However, doing so gives the model an obvious shortcut solution, which is to generate only the same features regardless of the input image. Thus, the loss of the network is always zero, and the learning of the network collapses. Therefore, negative samples are needed as a constraint, requiring the network to not only keep positive samples as close as possible but also keep negative samples as far away as possible. However, in BYOL, model collapse does not occur even without the negative samples, so it provides a new approach to contrastive learning.

BYOL does this by adding a predictor layer consisting of linear layers behind the projector of the query encoder. Let the output of the predictor predict the output of the projector layer of the key encoder. The original matching task of two views is turned into a prediction task that uses one view to predict the other. The query encoder updates using gradient backpropagation, while the key encoder does not update via gradient backpropagation but uses the parameters of the query encoder for momentum updates.

Xinlei Chen et al. proposed MoCo v3 in 2021 \cite{40}, which is the main reference for the contrastive learning method in this paper. MoCo v3 is no longer a simple improvement of MoCo v2 but borrows from BYOL. The predictor is also added at the end of the query encoder, replacing the pretext task with a prediction task, thus removing the negative sample restriction. Once the network does not need negative samples, it naturally does not need the queue structure to store them. However, the momentum encoder is retained, the query encoder is updated using gradient backpropagation, and the key encoder still uses momentum updates. And both encoders contain the projector proposed in SimCLR.

MoCo v3 uses the InfoNCE loss function commonly used for contrastive learning, as shown in (2).

\begin{equation}
\label{equ_2}
{{\cal L}_q} =  - \log \frac{{\exp \left( {q \cdot {k^ + }/\tau } \right)}}{{\exp \left( {q \cdot {k^ + }/\tau } \right) + \sum\limits_{{k^ - }} {\exp } \left( {q \cdot {k^ - }/\tau } \right)}}
\end{equation}

During the training process, let both data enhancement results ${x_1}$ and ${x_2}$ of $x$ go through query encoder and key encoder to get ${q_1}$, ${q_2}$, ${k_1}$, and ${k_2}$.When calculating the InfoNCE loss, the symmetric bidirectional loss is calculated as shown in (3).

\begin{equation}
\label{equ_3}
Loss = {\mathop{\rm InfoNCE}\nolimits} ({q_1},{k_2}) + {\mathop{\rm InfoNCE}\nolimits} ({q_2},{k_1})
\end{equation}

With the success of Transformer, there is a desire to apply Transformer to the CV field. So, Alexey Dosovitskiy et al. proposed ViT in 2020 \cite{41}. By splitting the image into small patches, Transformer has been successfully applied to vision tasks with fabulous results. Therefore, in MoCo v3, which was published after that, the backbone network was also changed from the previously used convolutional neural networks such as ResNet to ViT.

The experiments in this paper use a contrastive learning network with MoCo v3 as the baseline and make some adjustments on top of it. The specific adjustments will be explained in detail in Section III.

\section{PROPOSED METHOD}
\subsection{Contrastive learning method}
The contrastive learning method used in this paper is obtained by using MoCo v3 as the baseline improvement. The structure diagram of the contrastive learning network is shown in Fig. 1. The reason we chose to use MoCo v3 instead of a contrastive learning method such as MoCo v2 that uses instance discrimination as a pretext task is that the instance discrimination task considers each image as a separate category. Images between different categories form negative sample pairs with each other, and the goal of contrastive learning is to keep the negative sample pairs away from each other in the feature space. This means that even if the two images are very similar, the encoder will try to distinguish them as much as possible.

\begin{figure}[!t]
\centering
\includegraphics[width=\columnwidth]{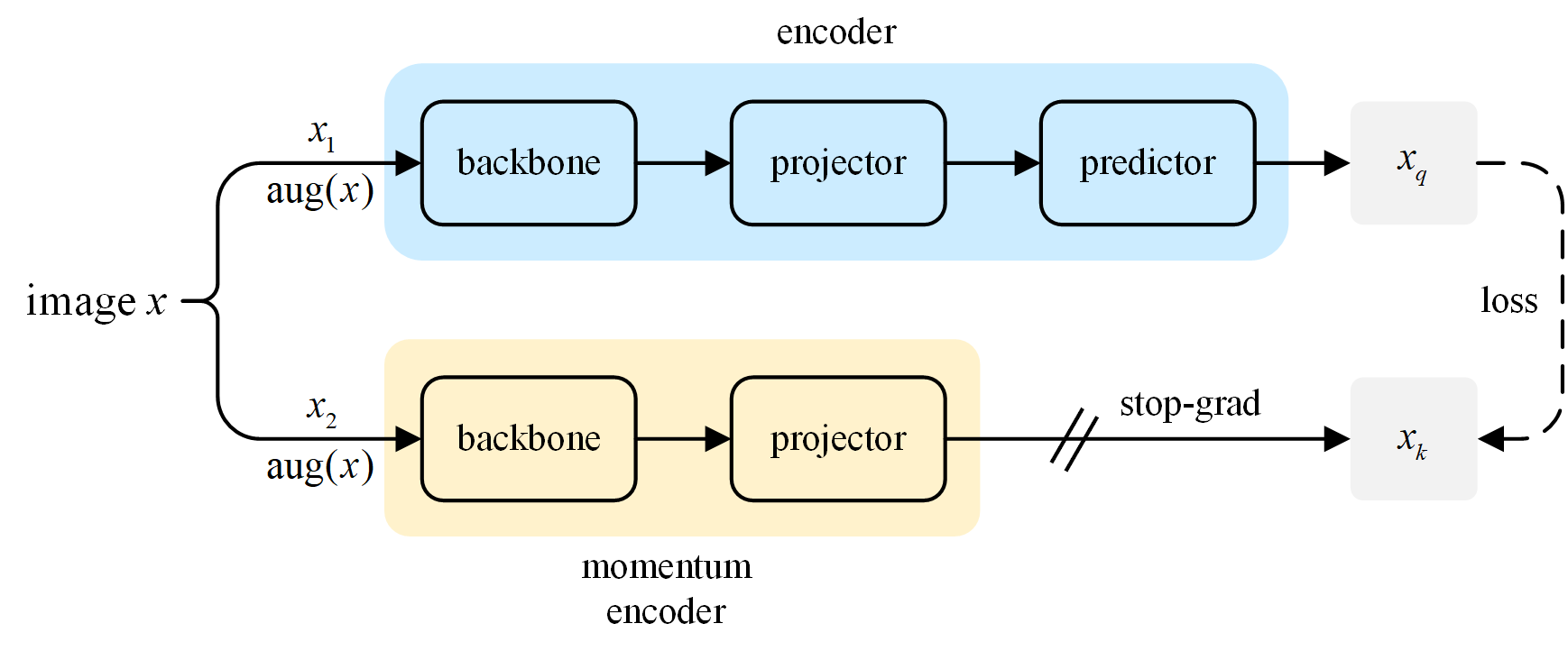}
\caption{Structure diagram of the contrastive learning network.}
\label{fig_1}
\end{figure}

In natural image datasets, most images are very different and rarely have similar images, so the instance discrimination task is reasonable. For medical imaging, however, a single CT or MRI examination produces multiple slices that are very similar to each other. Even with image data from different patients, there is often a high degree of similarity between images due to the consistency of the body structure. So, in medical image datasets, there are many similar images that make it unsuitable to use instance discrimination as a pretext task.

The pretext task of MoCo v3 is no longer the previous matching task, and the training of the network no longer relies on the negative sample constraint. Therefore, it does not have the problem of being unsuitable for the instance discrimination task due to too much similarity between medical images. Although MoCo v3 requires a larger batch size than MoCo v2, we still use MoCo v3 as the baseline to make changes, as follows.

First, the backbone network of MoCo v3 was replaced. Since the segmentation network used for the downstream segmentation task is based on Swin-Unet, whose backbone network is Swin Transformer. Therefore, the backbone network of MoCo v3 needs to be replaced from ViT to Swin Transformer so that the encoder model trained by contrastive learning can be used as a pre-trained model for Swin-Unet transfer learning.

Second, the data enhancement method was changed. The data enhancement methods used in the original MoCo v3 paper include random cropping, color dithering, random grayscale, Gaussian blur, random Solarize, and random horizontal flipping. In general, more data enhancement is more beneficial for the contrastive learning model to learn the key semantic information of the images. Therefore, according to the characteristics of parotid MR images, we added random small-angle rotation at the beginning to simulate the small-angle head rotation that patients may have during MRI examinations.

Third, the hyperparameters of the training process are adjusted. The batch size used in the original MoCo v3 is 4096, and the ImageNet dataset used contains 1.28 million natural images in the train set. Using a large batch size on such a large dataset will bring better results. However, due to the characteristics of medical images, the dataset size will be much smaller. The parotid dataset used in this paper has 4688 images that can be used for contrastive learning after preprocessing, and a batch size as large as 4096 cannot be used, so we set the batch size to 96. In addition, due to the reduction in the amount of data, the number of steps contained in each epoch during network training becomes less, and correspondingly the speed of network gradient descent must be increased. Therefore, the learning rate maximum increased from $1.5 \times {10^{ - 4}}$ to $1 \times {10^{ - 3}}$ by grid search experiments.

\subsection{Using transfer learning to train the contrastive learning network}
Contrastive learning, as an unsupervised learning method, has been compared to supervised learning methods in previous work, including in the transfer of downstream tasks. In the MoCo v1 paper, the results of using contrastive learning for classification tasks have approached the level of supervised learning. In other mainstream CV tasks, it even surpasses the results of supervised learning. However, we do not train the contrastive learning model from scratch using a random initialization parameter approach. Instead, we use the pre-trained supervised learning model on ImageNet as the initialization parameters for the contrastive learning model, in other words, using transfer learning to train the contrastive learning network. There are two main reasons for doing so.

On the one hand, as an unsupervised learning method, contrastive learning does not require the necessary labels in supervised learning. Therefore, to achieve equal or better results, the requirements for data set size are more demanding. The commonly used datasets in previous studies are ImageNet, COCO, and even larger datasets than them. However, due to the characteristics of medical images, the relatively large medical image datasets are only about a few thousand cases, which cannot match the size of natural image datasets. Therefore, directly training the contrastive learning network with random initialization is not effective.

On the other hand, with the recent development of Transformer, researchers have started to experiment with replacing the backbone network of contrastive learning with Transformer, which can be more demanding on the dataset. Convolutional neural networks, which were designed and proposed from the beginning to deal with image problems, have a lot of priori knowledge about images. For example, the convolution operation causes the situation that network can get the same features from the same images in different locations, which leads to the assumption of feature shift-invariance. Similar assumptions that have been introduced at the time of designing the method are what we call inductive bias. However, Transformer, an architecture first used in natural language processing, has the advantage of global modeling over convolutional neural networks but lacks inductive bias for images. Therefore, a large amount of data is needed for pre-training to compensate for this deficiency. But the dataset of medical images is not large enough for the network to learn these missing inductive biases. Therefore, the contrastive learning network trained in this way may not perform as well as the supervised learning network.

In summary, to overcome the problem that a small dataset cannot train a good contrastive learning model, it is necessary to use transfer learning to train the contrastive learning network.

\subsection{Segmentation network}
The segmentation network used in this paper is based on Swin-Unet, which is obtained by replacing the backbone network of U-Net with Swin Transformer. The encoder and bottleneck parts together are a tiny version of Swin Transformer. The decoder part contains the Swin Transformer Block layers corresponding to the encoder, but also adds the Patch Expanding layers as the upsampling layers. The Linear Projection head is used at the end of the network to implement the mapping from features to segmentation results. Skip Connection is used as U-Net to add low-dimensional information in the encoder to the decoder. The structure of the network is clear and simple, yet it can achieve better results than traditional convolutional neural networks. 

Based on the experimental findings of the previous paper, this paper still uses multimodal data to form three channels and mixes the multicenter data before randomly selecting the train and test sets. The difference is that this paper makes some adjustments to the training method to ensure that the performance of the network is fully exploited, as follows.

First, the hyperparameters used for training were adjusted. A larger batch size generally leads to better training results, so the batch size is changed from the original 8 to the current 16. The learning rate drop method is also changed. The original learning rate is fixed to $1 \times {10^{ - 4}}$ during the training process, which is not conducive to model convergence. Therefore, the learning rate is changed to decrease as a half-cycle cosine function during training, as shown in (4). Where $lr$ is the learning rate used for the current epoch, $l{r_{\max }}$ is the maximum learning rate, $i$ is the current number of epochs, and $m$ is the total number of epochs. And due to the decrease in the number of training steps caused by the increase in batch size, the maximum value of the learning rate is increased to $2 \times {10^{ - 4}}$.

\begin{equation}
\label{equ_4}
lr = l{r_{max}} \cdot (1 + \cos (\frac{i}{m} \cdot \pi ))/2
\end{equation}

Second, the data enhancement method is added. Due to the small size of the dataset, the network is prone to overfitting during the training process, resulting in degraded performance on the test set. Therefore, we used some data enhancement methods on the train set to mitigate overfitting, including color dithering, Gaussian blurring, and small angle rotation. After adding data enhancement, the overfitting problem of training is alleviated to some extent.

It should be noted that the data enhancement for Swin-Unet does not use random cropping, and the black edges resulting from the rotation are not cropped after the small-angle rotation operation. Because a big difference between medical images and natural images is that medical images are obtained by using specific examination instruments under the operation of professional imaging doctors. So there is no cropping of some objects, which is common in natural images. Therefore, introducing cropping operations in the data enhancement step can destroy the integrity of medical images, which leads to worse training results.

\subsection{Collected Dataset}
The raw data used in this paper are consistent with those in our previous paper, with multimodal MR images collected from a total of 148 patients with parotid tumors from two different centers. The raw data contain MRI sequences of three modalities, namely short time inversion recovery (STIR) sequences, T1-weighted sequences, and T2-weighted sequences. Experienced clinicians select the side with the tumor from both sides of the parotid gland and outline the ROI of the parotid gland and the tumor separately as segmentation labels.

The dataset used to train the segmentation network is consistent with that in the previous paper. The STIR sequences, T1-weighted sequences, and T2-weighted sequences of the image are composed into three-channel images. A total of 1897 MR image slices were available after pre-processing. 80\% of the dataset is taken as the train set and the remaining 20\% as the test set. Some images of the dataset used to train the segmentation network are shown in Fig. 2. The color image in the figure is the effect of a three-channel image consisting of STIR, T1, and T2 displayed in RGB format.

\begin{figure}[!t]
\centering
\includegraphics[width=\columnwidth]{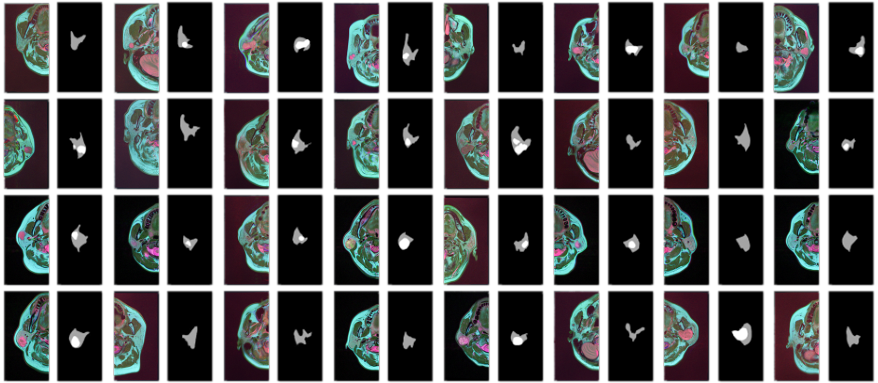}
\caption{Some images of the segmentation dataset.}
\label{fig_2}
\end{figure}

The datasets used to train the contrastive learning network and the segmentation network in this paper are different. Contrastive learning, as an unsupervised learning method, does not rely on labels for its training process. Therefore, unlabeled images can also be used for contrastive learning training, which achieves the purpose of putting the data to good use. The data set for training contrastive learning is also composed of three channel images with STIR, T1, and T2, and a total of 4688 images are available after pre-processing. Some images of the dataset used for training contrastive learning are shown in Fig. 3.

\begin{figure}[!t]
\centering
\includegraphics[width=\columnwidth]{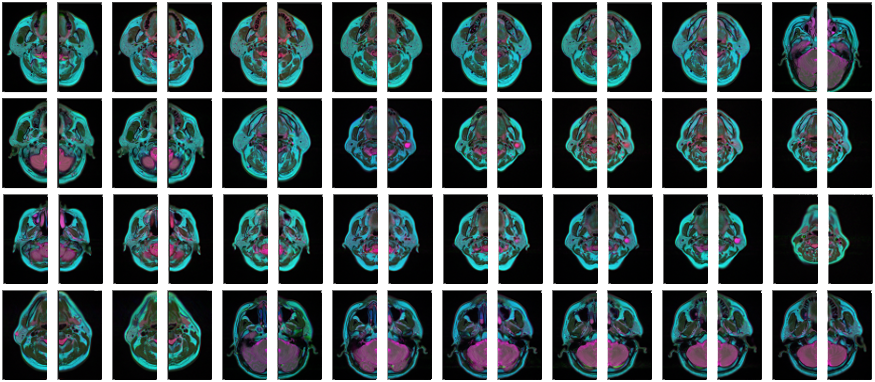}
\caption{Some images of the contrastive learning dataset.}
\label{fig_3}
\end{figure}

\section{EXPERIMENTS AND DISCUSSION}
In this section, we designed three different ablation studies. The contrastive learning network and the segmentation network used were set up according to Section III, except for the experimental setup that was specifically described. As in the previous paper, this paper continues to use four segmentation evaluation metrics to evaluate the segmentation effectiveness of the model on the test set. They are Dice-Similarity coefficient (DSC), Mean Pixel Accuracy (MPA), Mean Intersection over Union (MIoU), and Hausdorff Distance (HD), respectively. The formulas for the four metrics are shown in (5), (6), (7), and (8), where $h\left( {A,B} \right)$ and $h\left( {B,A} \right)$ in equation (8) are calculated as shown in (9) and (10).

\begin{equation}
\label{equ_5}
{\rm{ DSC }} = \frac{{2TP}}{{FP + 2TP + FN}}
\end{equation}

\begin{equation}
\label{equ_6}
MPA = \frac{{TP + TN}}{{FN + TP + FP + TN}}
\end{equation}

\begin{equation}
\label{equ_7}
MIoU = \frac{{TP}}{{FN + TP + FP}}
\end{equation}

\begin{equation}
\label{equ_8}
H(A,B) = \max (h(A,B),h(B,A))
\end{equation}

\begin{equation}
\label{equ_9}
h(A,B) = \max (a \in A)\min (b \in B)\quad \left\| {a - b} \right\|
\end{equation}

\begin{equation}
\label{equ_10}
h(B,A) = \max (b \in B)\min (a \in A)\quad \left\| {b - a} \right\|
\end{equation}

\subsection{Experiments to validate the effectiveness of training the contrastive learning network with transfer learning}
To verify the effectiveness of training the contrastive learning network with transfer learning, an ablation study containing four experiments was designed. The four experiments used the same segmentation network, differing only in that the segmentation network used different pre-trained models for parameter initialization. The first experiment does not use a pre-trained model and directly initializes the parameters randomly for training. The second experiment uses a supervised learning model pre-trained on ImageNet. The third experiment uses a pre-trained model of the contrastive learning network, which is trained with random initialization parameters. The fourth experiment also uses a pre-trained model of the contrastive learning network, but this time the contrastive learning network is trained with a supervised learning model as the initialization parameter, which is equivalent to training the contrastive learning model with transfer learning. Since the fourth experiment is relatively complex, the training flow is shown in Fig. 4. for ease of understanding.

\begin{figure}[!t]
\centering
\includegraphics[width=\columnwidth]{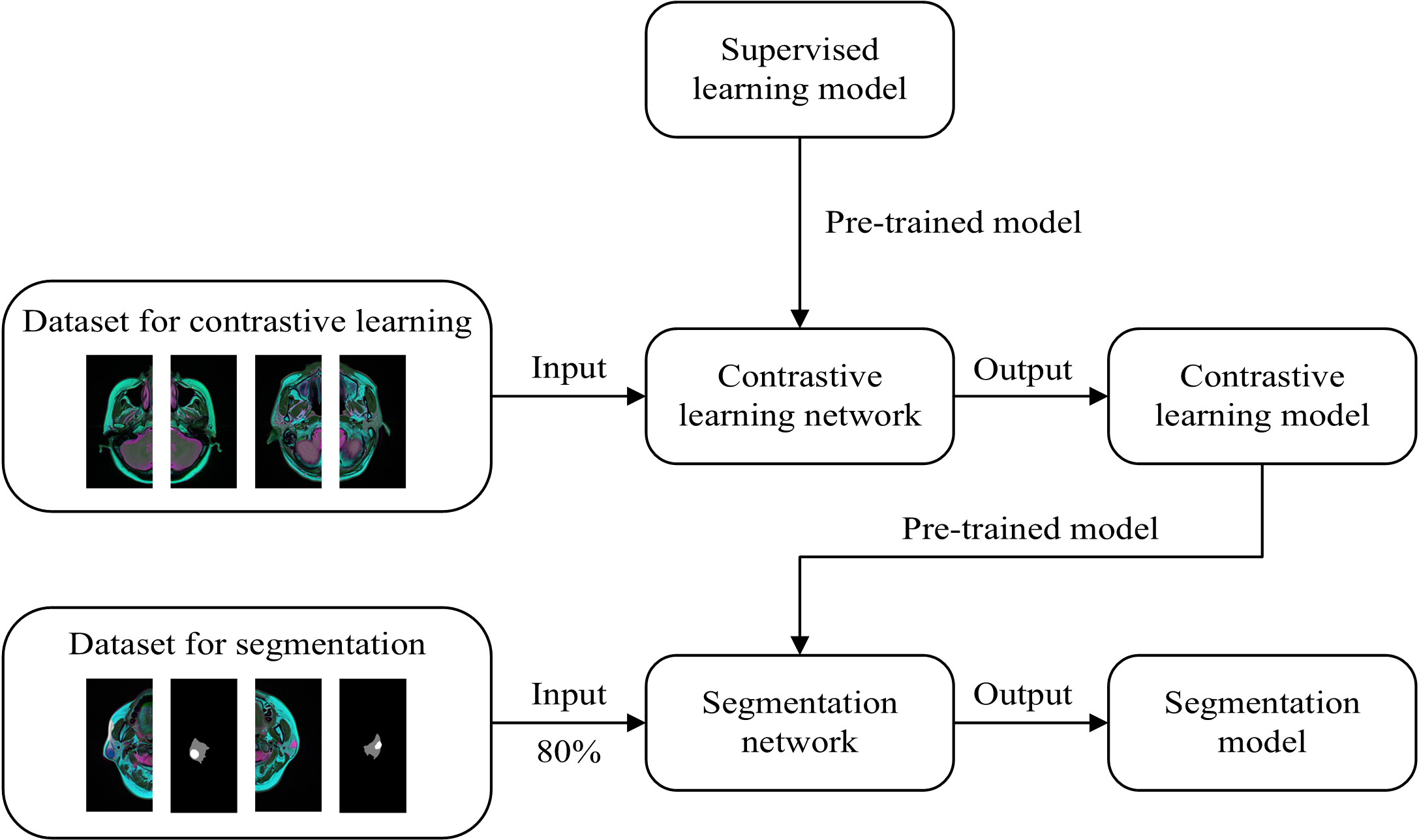}
\caption{Schematic diagram of the training process for the fourth experiment.}
\label{fig_4}
\end{figure}

The results of the four experiments are as follows. The curve of the loss function on the test set during the training process of the segmentation network is shown in Fig. 5. The evaluation metrics of the model on the test set are shown in Table I.

\begin{figure}[!t]
\centering
\includegraphics[width=\columnwidth]{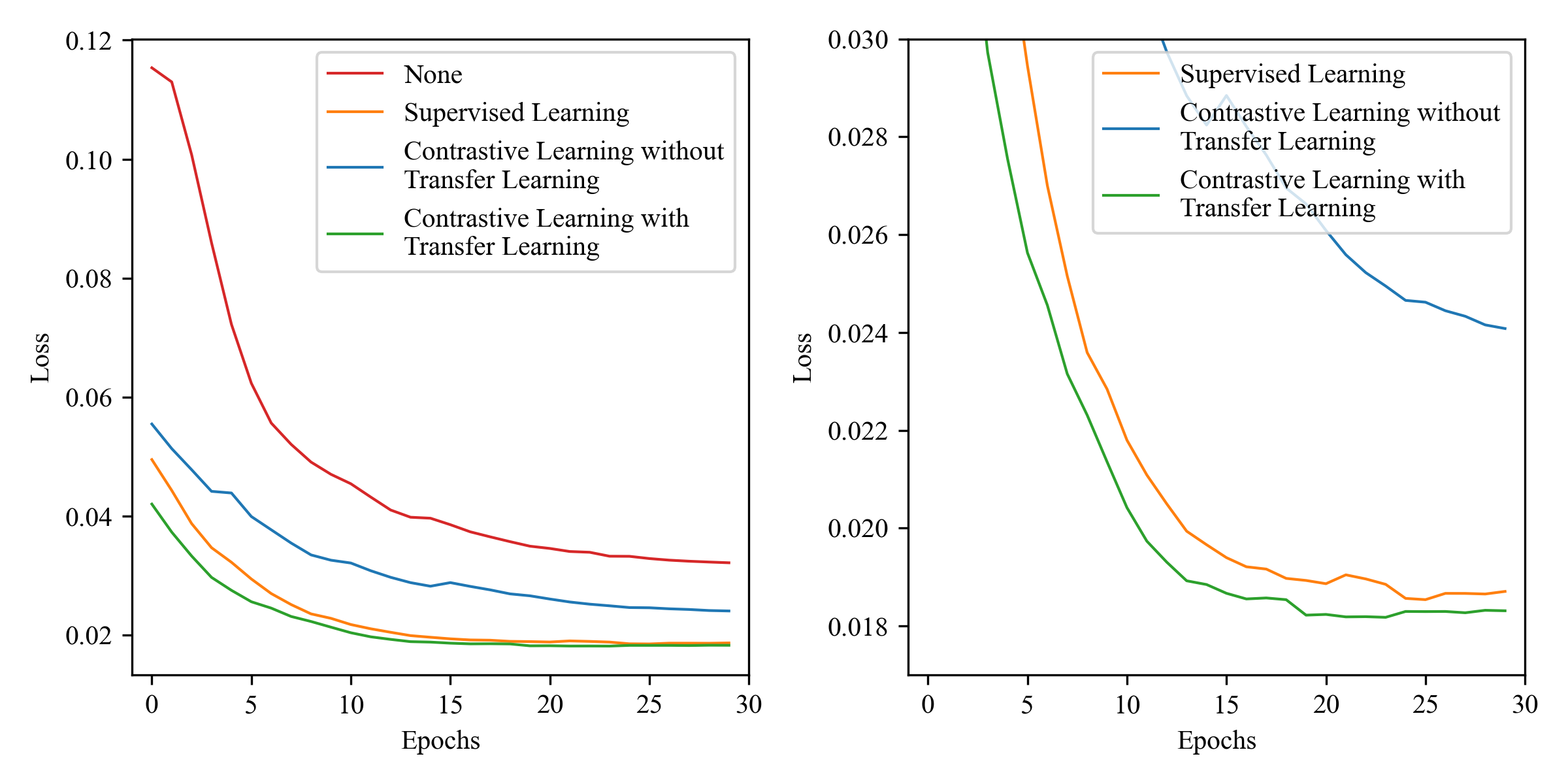}
\caption{Curves of the loss function on the test set for the first ablation study. The two graphs show different ranges of vertical axes.}
\label{fig_5}
\end{figure}

\begin{table}
\caption{Evaluation metrics for the models of\\the first ablation study on the test set.}
\label{tab_1}
\resizebox{\columnwidth}{!}{
\begin{tabular}{r c c c c}
\hline
Pre-trained model & DSC(\%) & MPA(\%) & MIoU(\%) & HD\\
\hline
None & 75.22 & 98.82 & 69.23 & 3.82\\
Supervised Learning & 88.52 & 99.31 & 83.89 & 3.06\\ 
{\makecell[r]{Contrastive Learning without\\Transfer Learning}} & 83.95 & 99.08 & 78.51 & 3.37\\ 
\textbf{\makecell[r]{Contrastive Learning with\\Transfer Learning}} & \textbf{89.60} & \textbf{99.36} & \textbf{85.11} & \textbf{2.98}\\  
\hline
\end{tabular}
}
\end{table}

Based on the experimental results, it is known that the contrastive learning model trained with random initialization parameters has some effect. Using this model as a pre-trained model for the segmentation network outperforms the results of not using the pre-trained model. But as mentioned in Section III, Transformer is missing some important inductive biases. And the dataset size is not sufficient for the network to learn these inductive biases from it. Therefore, the final performance is not as good as the results of using the supervised learning model as a pre-trained model for the segmentation network. However, the best model results are obtained when the contrastive learning network is trained with transfer learning, and this result can be explained by the following three reasons.

First, the pre-trained model with supervised learning contains these missing inductive biases. Using transfer learning allows introducing them at the beginning of training the contrastive learning network, thus solving the above problem. Second, the supervised learning model is trained on the R, G, and B three-channel natural image dataset, which is quite different from the STIR, T1, and T2 three-channel medical image dataset used in this experiment. In comparison, the models obtained by contrastive learning are trained on these medical images, and the higher similarity of the data domains naturally leads to better transfer learning results. Third, only one side of the parotid gland was labeled in the original data used in this paper, which means that the segmentation network can only use up to half of the image data for training. Contrastive learning does not rely on labels and can be trained using the full range of image data. The increase in the amount of data allows the network to learn better features from more samples, thus improving the performance of the segmentation model.

\subsection{Experiments to verify that the contrastive learning model is valid only for the encoder of the segmentation network}
In the original paper of Swin-Unet, the authors used a pre-trained model with supervised learning to do transfer learning. Since the encoder and bottleneck parts of Swin-Unet together are the Swin Transformer network, the pre-trained model of Swin Transformer can easily be used as the initialization weights for the encoder and bottleneck parts. In the decoder part, the Patch Merging layers are replaced with Patch Expanding layers, so only the Swin Transformer Block layers in the decoder use the same weights as the corresponding layers in the encoder. The remaining Patch Expanding layers, Skip Connection layers and Linear Projection layer are trained with random initialization parameters. Our previous paper and the previous ablation study in this paper both did transfer learning of segmentation networks in this way. 

However, the contrastive learning method used in this paper is to obtain images that look different but have the same semantic information by different ways of data enhancement, thus enabling the encoder to learn deeper features. In other words, the encoder must be trained to be resistant to interference, so that the encoder will give the same features no matter how the image is enhanced. Such a training logic is built for an encoder. Then it is still an open question whether the improvement brought by the contrastive learning model over the supervised learning model still holds in the decoder part. Therefore, we designed the following ablation study to investigate this problem. For the convenience of discussion, the encoder in the following contains both encoder and bottleneck, and the contrastive learning model refers to the model obtained by training the contrastive learning network with transfer learning.

We divide the segmentation network into two parts: the encoder and the decoder. Each part can choose three training methods, namely, no pre-trained model, using a supervised learning model as a pre-trained model, and using a contrastive learning model as a pre-trained model. This constitutes a total of nine experiments. The experimental results are as follows. The curves of the loss function on the test set during the training of the network are shown in Fig. 6. The evaluation metrics of the model on the test set are shown in Table II. The numbers in the figure legend names are the experiment serial numbers. Before the underline represents the model used by the encoder and after the underline represents the model used by the decoder. None means no pre-trained model is used, Supe means the supervised learning model is used, and Cont means the contrastive learning model is used.

\begin{figure}[!t]
\centering
\includegraphics[width=\columnwidth]{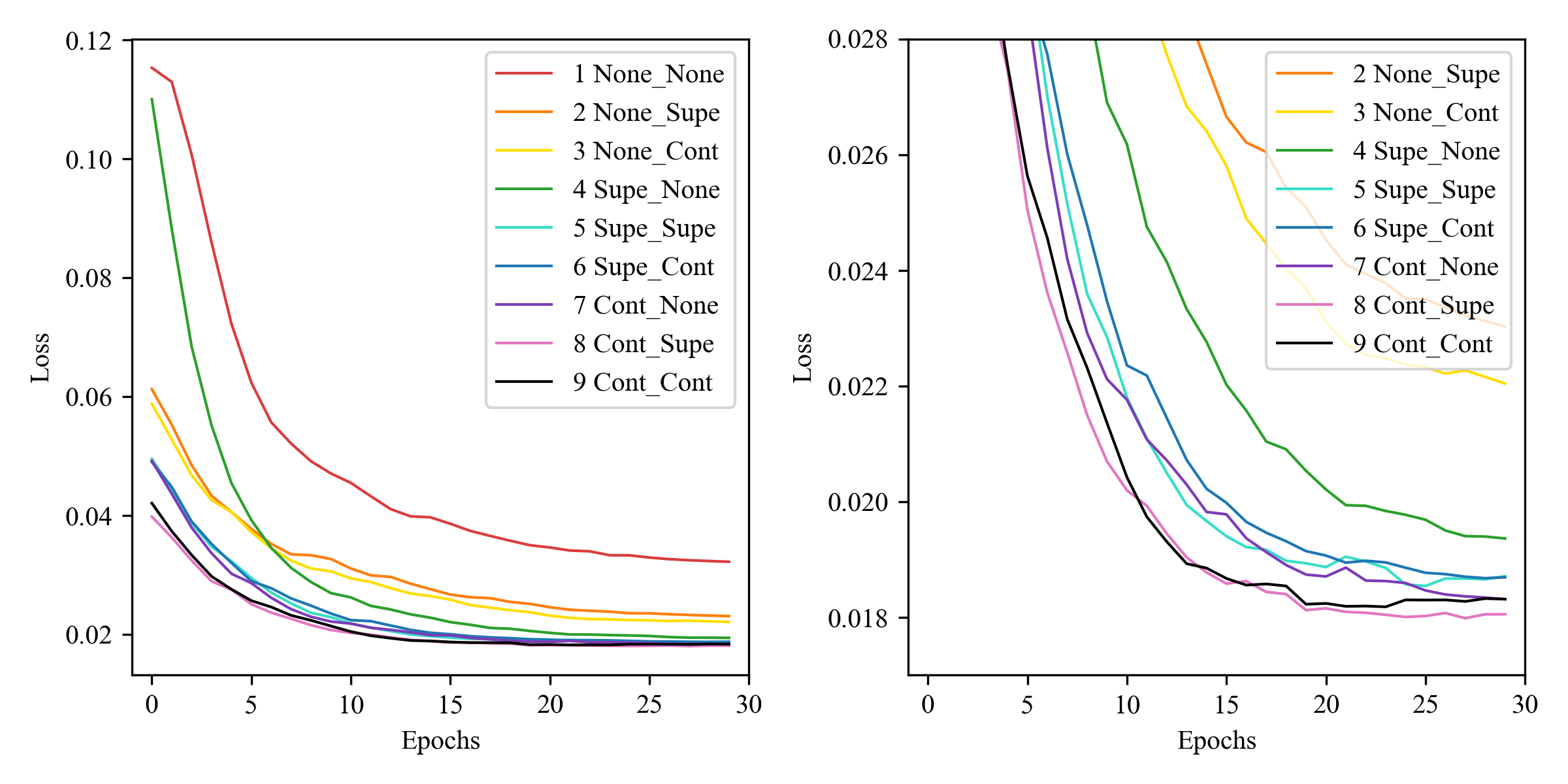}
\caption{Curves of the loss function on the test set for the second ablation study. The two graphs show different ranges of vertical axes.}
\label{fig_6}
\end{figure}

\begin{table}
\caption{Evaluation metrics for the models of\\the second ablation study on the test set.}
\label{tab_2}
\resizebox{\columnwidth}{!}{
\begin{tabular}{c c c c c c c}
\hline
Index & Encoder & Decoder & DSC(\%) & MPA(\%) & MIoU(\%) & HD\\
\hline
1 & None & None & 75.22 & 98.82 & 69.23 & 3.82 \\
2 & None & Supe & 85.04 & 99.13 & 79.75 & 3.28 \\
3 & None & Cont & 85.25 & 99.17 & 80.08 & 3.26 \\
4 & Supe & None & 87.32 & 99.28 & 82.45 & 3.11 \\
5 & Supe & Supe & 88.52 & 99.31 & 83.89 & 3.06 \\
6 & Supe & Cont & 88.22 & 99.31 & 83.53 & 3.05 \\
7 & Cont & None & 88.19 & 99.33 & 83.51 & 3.04 \\
\textbf{8} & \textbf{Cont} & \textbf{Supe} & \textbf{89.95} & \textbf{99.36} & \textbf{85.47} & \textbf{3.00} \\
\textbf{9} & \textbf{Cont} & \textbf{Cont} & \textbf{89.60} & \textbf{99.36} & \textbf{85.11} & \textbf{2.98} \\
\hline
\end{tabular}
}
\end{table}

By comparing the first three experiments, the middle three experiments, and the last three experiments, it can be found that when the encoder settings are fixed, the results obtained by the decoder using the supervised learning model or using the contrastive learning model are almost the same, and both are better than the results obtained by the decoder without the pre-trained model. This illustrates that the contrastive learning model does not improve the decoder significantly compared to the supervised learning model.

In addition, the best results among the nine experiments were in Experiment 8 and Experiment 9, where the encoder part used the contrastive learning model. Experiment 7 with the encoder using the contrastive learning model even though the decoder did not use the pre-trained model, reached the same level as Experiments 5 and 6 with the encoder using the supervised learning model and the decoder using a pre-trained model. This further illustrates that the contrastive learning model improves the encoder very significantly compared to the supervised learning model.

In summary, the experiments conclude that the contrastive learning model only improves the performance of the encoder and the change brought to the decoder is not significant. As mentioned in the theoretical analysis above, the contrastive learning method used in this paper is designed based on the behavior of the encoder. However, to our knowledge, almost all contrastive learning methods, including all those mentioned in Section II, are designed for the encoder.

\subsection{Experiments on contrastive learning networks for the decoder}
Due to the above, we try to design a contrastive learning network for the decoder and expect to further improve the performance of the segmentation model. For the encoder, the input is images, and the output is encoded features. Therefore, data enhancement can be performed on the input image and the encoder output features are required to be unchanged so that the encoder learns the semantic features in the image that are not prone to change. We tried to construct a contrastive learning network for the decoder using the same way. For the decoder, the input is the features output from the encoder and the output is the decoded features. Therefore, we make the output features of the encoder change by data enhancement and require the output of the decoder not to change, thus serving the purpose of training the decoder. The structure diagram of the designed network is shown in Fig. 7.

\begin{figure*}[!t]
\centering
\includegraphics[width=5in]{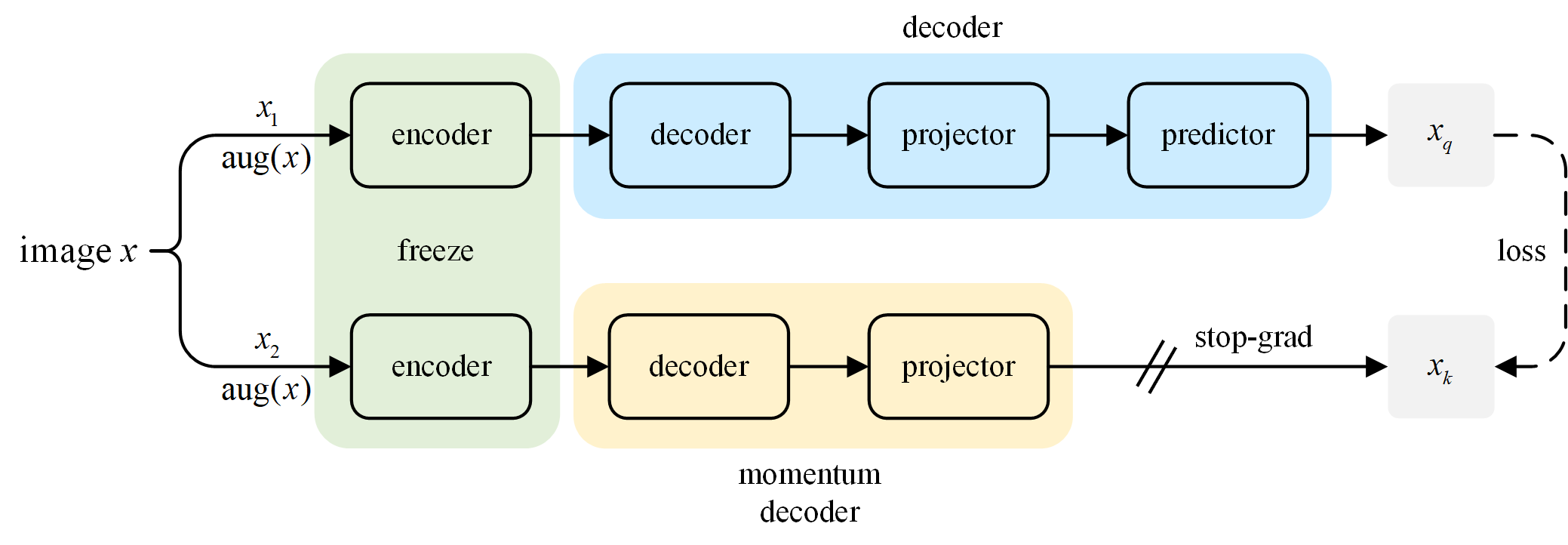}
\caption{Structure diagram of contrastive learning network for decoder.}
\label{fig_7}
\end{figure*}

To avoid the network training the encoder, the weights of the encoder are frozen during the training process. Instead of using gradient backpropagation update, momentum decoder uses the parameters of base decoder for momentum update. Skip Connection layers from Swin-Unet are also added between the encoder and decoder, which are not drawn in the figure. After the training is completed, only the decoder in the base decoder is retained and used as the initialization parameter for the decoder part of the segmentation network. However, the designed network was not as good as we expected after several adjustments. The following discussion is based on the ablation study during the adjustment process, the results of which are shown in Table III. The pre-trained models used for the encoders of these five experiments are all contrastive learning models, and the pre-trained models used for the decoder part are all models obtained from the contrastive learning network for the decoder. For comparison, the results of Experiment 9 from the previous ablation study have been placed in the first row of the table.

\begin{table}
\caption{Experiments on tuning contrastive learning network\\for the decoder}
\label{tab_3}
\resizebox{\columnwidth}{!}{
\begin{tabular}{c c c c c c}
\hline
Index & Training method & DSC(\%) & MPA(\%) & MIoU(\%) & HD\\
\hline
  & \textbf{None} & \textbf{89.60} & \textbf{99.36} & \textbf{85.11} & \textbf{2.98} \\
1 & Random initialization & 88.73 & 99.30 & 83.94 & 3.07 \\
2 & Contrastive learning & 88.51 & 99.30 & 83.79 & 3.05 \\
3 & Supervised learning & 	89.38 & 99.34 & 84.77 & 3.01 \\
4 & Change data enhancement & 88.55 & 99.31 & 83.85 & 3.06 \\
5 & Remove other parameters & 88.90 & 99.30 & 84.14 & 3.04 \\
\hline
\end{tabular}
}
\end{table}

The first problem we face is the selection of the initialization parameters of the encoder. The parameters of the encoder are frozen and do not change throughout the training process, thus ensuring that the network is trained against the decoder. Therefore, the selection of the initialization parameters of the encoder becomes particularly important. We initially train the entire contrastive learning network for the decoder with random initialization parameters, and the results are shown in Experiment 1. But we soon discover that there was a major problem with doing so. Since the encoder weights are randomly initialized, the encoder has no encoding capability. It outputs only meaningless vectors after random operations, not the image features output by the encoder. Therefore, it is not possible to train the decoder effectively.

But if the initialization parameter used by the encoder is the contrastive learning model, this leads to the other extreme. The results are shown in Experiment 2. Since the encoder of the contrastive learning model has already been resistant to data enhancement of images, the features of its output has already hardly changed with data enhancement. But in this case, decoder will only gets nearly the same input all the time. Thus, it also can not be trained effectively.

Therefore, we can only choose to use the pre-trained model with supervised learning as the initialization parameters of the encoder, and the results are shown in Experiment 3. The results of this experiment exceed those of Experiment 1 and Experiment 2 but still do not exceed the best results of the previous experiments.

Next, we tried to change the data enhancement method based on Experiment 3, and the results are shown in Experiment 4. The previous data enhancement includes random cropping and small angle rotation. The images after data enhancement will produce different segmentation results when they are input to the segmentation network. However, the decoder output features are already very close to the segmentation result. Since the decoder output is required not to change with data enhancement, those data enhancement methods should be chosen that do not necessarily affect the segmentation result. Here we choose color dithering and Gaussian blur. Surprisingly, the results go down a lot this time instead. One possible explanation for this is that contrastive learning networks are more dependent on data augmentation, and more data augmentation tends to lead to better results, just as SimCLR uses more data augmentation to make the results better. However, the removal of random cropping and small-angle rotation here actually weakens the effect of data enhancement and prevents the network from learning good features from it.

Using contrastive learning for the decoder can solve another problem. As mentioned above only the Swin Transformer Block layers of the decoder can use the pre-trained model. However, the entire decoder part is now being trained with contrastive learning, which includes the Patch Expanding layers and the Skip Connection layers, thus solving the problem that these layers do not have pre-trained model parameters. But none of the four experiments from Experiment 1 to Experiment 4 exceeded the previous best results. To verify whether the use of pre-trained models for the Patch Expanding layers and Skip Connection layers leads to a boost, we tried to build on Experiment 3 by removing the pre-trained parameters of the Patch Expanding layers and Skip Connection layers and letting them initialize randomly for training. The results are shown in Experiment 5. It can be found that experiment 5 has a significant decrease in accuracy compared with experiment 3, which illustrates that the pre-trained parameters of Patch Expanding layers and Skip Connection layers are effective. There is hope that the performance of the segmentation model can be further better if the contrastive learning network for the decoder can be improved. Therefore, the improvement of the contrastive learning network for the decoder will be our next research objective.

\section{CONCLUSION}
In this paper, we used a Transformer-based contrastive learning method and innovatively trained the contrastive learning network with transfer learning. Then, the output model was transferred to the downstream parotid segmentation task, which improved the performance of the parotid segmentation model on the test set. The improved DSC was 89.60\%, MPA was 99.36\%, MIoU was 85.11\%, and HD was 2.98. All four metrics show significant improvement compared to the results of using a supervised learning model as a pre-trained model for the segmentation network. Better segmentation results are more useful to assist clinicians in the diagnosis and treatment planning of PGTs. Objective, accurate and rapid segmentation methods can also facilitate researchers to conduct more research on PGTs.

Further, we consider that not only the parotid gland segmentation task but also any task with a small dataset and relying on transfer learning can be attempted by replacing the used pre-trained model from a supervised learning model with a contrastive learning model trained with transfer learning. Because training the contrastive learning network with transfer learning introduces the knowledge already learned by the supervised learning model, which contains the inductive bias missing in the Transformer network. Therefore, the problem of small datasets not being able to meet the training needs of contrastive learning networks is alleviated to some extent.

In addition, we also find that for segmentation networks with both encoder and decoder, the pre-trained model obtained with contrastive learning is improved mainly for the encoder part, while there is no significant improvement for the decoder part. Subsequently, an attempt was also made to construct a contrastive learning network for the decoder, but after several adjustments, it still did not achieve the expected results. We consider that it would be helpful to use contrastive learning for the decoder to obtain a better pre-trained model as the initialization parameter for the decoder, which would help to make the performance of the segmentation model further improved.

\bibliographystyle{IEEEtran}
\bibliography{paper}

\clearpage

\begin{IEEEbiography}[{\includegraphics[width=1in,height=1.25in,clip,keepaspectratio]{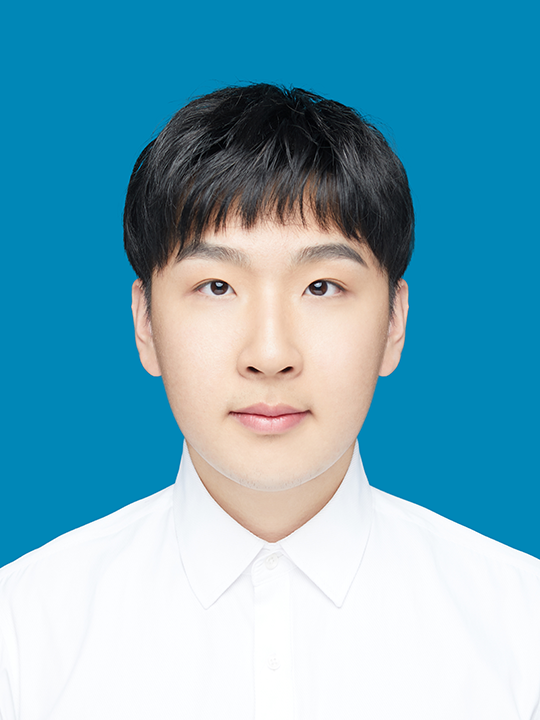}}]{Zi'an Xu}
is currently pursuing his master's degree in College of Medicine and Biological Information Engineering from Northeastern University, Shenyang, China. His research interests focus on deep learning in computer vision, classification and segmentation of medical images, and computer-aided diagnosis.
\end{IEEEbiography}

\vspace{11pt}

\begin{IEEEbiography}[{\includegraphics[width=1in,height=1.25in,clip,keepaspectratio]{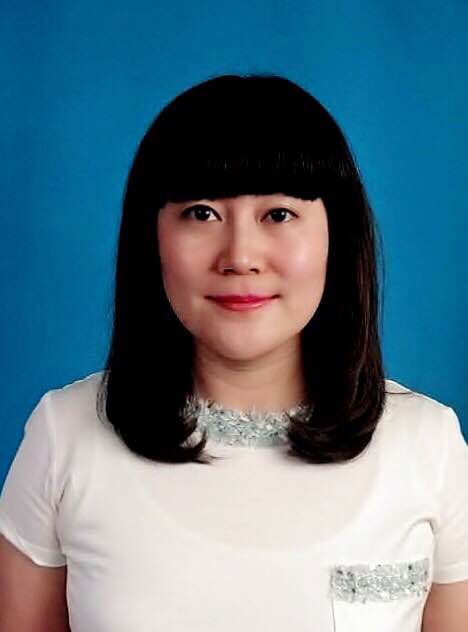}}]{Yin Dai}
received the Ph.D. degree from the Department of Computer Science, Northeastern University, China, in 2015. She is currently an associate professor in the College of Medicine and Biological Information Engineering at Northeastern University, China. Her research lies at computer-aided diagnosis and medical image processing.
\end{IEEEbiography}

\vspace{11pt}

\begin{IEEEbiography}[{\includegraphics[width=1in,height=1.25in,clip,keepaspectratio]{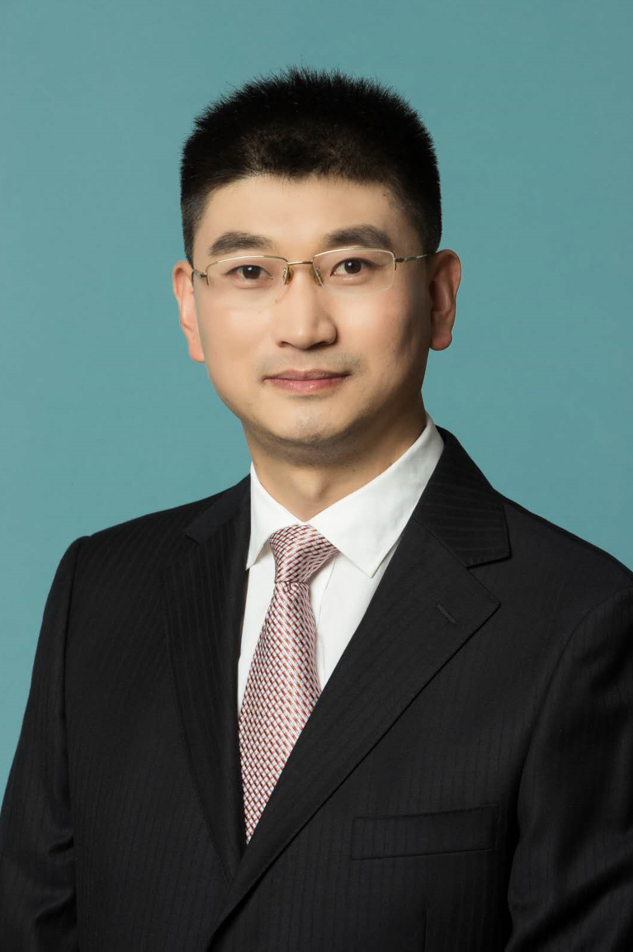}}]{Fayu Liu}
Professor, Chief Physician, Doctoral Supervisor, Director of the Department of Oral and Maxillofacial Surgery, Oral Hospital of China Medical University. He is a member of the Standing Committee of Head and Neck Specialty Committee of China Medical Education Association; a member of Skull Base Surgery Branch of China Association for the Promotion of Healthcare International Exchange; a youth member of Oral and Maxillofacial-Head and Neck Tumor Specialty Committee of Chinese Oral Medical Association; a youth member of Head and Neck Tumor Specialty Committee of Chinese Anti-Cancer Association. He has been engaged in clinical, teaching and research work in oral and maxillofacial-head and neck surgery. He has received a Research Fellowship from Sloan-Kettering Cancer Center, Rosewell Park Cancer Center and the State University of New York at Buffalo, and has led one National Natural Science Foundation, four provincial-level projects, two municipal-level projects, and participated in one national key research and development project. He has published 20 papers in foreign SCI journals and many papers in domestic core journals as the first author or corresponding author. He participated in writing the book "Repair and Reconstruction of Head and Neck Defects" published by People's Health Publishing House, and won one second prize and two third prizes of Science and Technology Progress of Liaoning Provincial Government.
\end{IEEEbiography}

\vspace{11pt}

\begin{IEEEbiography}[{\includegraphics[width=1in,height=1.25in,clip,keepaspectratio]{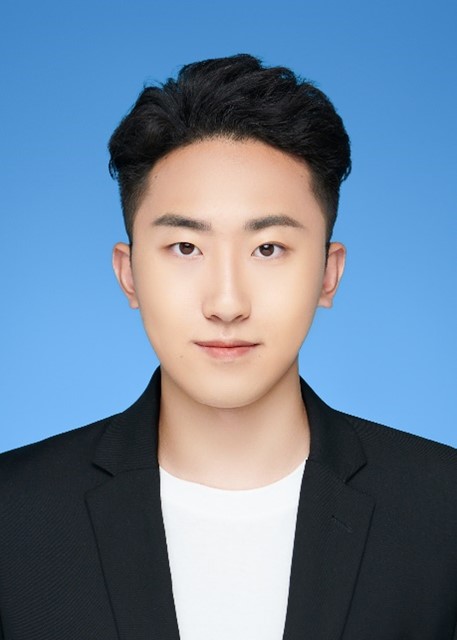}}]{Boyuan Wu}
is an undergraduate student in the College of Medicine and Biological Information Engineering at Northeast University, China. His research lies at computer-aided diagnosis and medical image processing.
\end{IEEEbiography}

\vspace{11pt}

\begin{IEEEbiography}[{\includegraphics[width=1in,height=1.25in,clip,keepaspectratio]{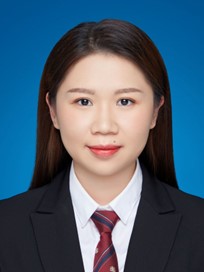}}]{Weibing Chen}
received the B.E. degree from the College of Medicine and Biological Information Engineering , Northeastern University, China, in 2022. She is a current graduate student in the College of Medicine and Biological Information Engineering at Northeastern University, China. Her research lies at medical image big data processing.
\end{IEEEbiography}

\vspace{11pt}

\begin{IEEEbiography}[{\includegraphics[width=1in,height=1.25in,clip,keepaspectratio]{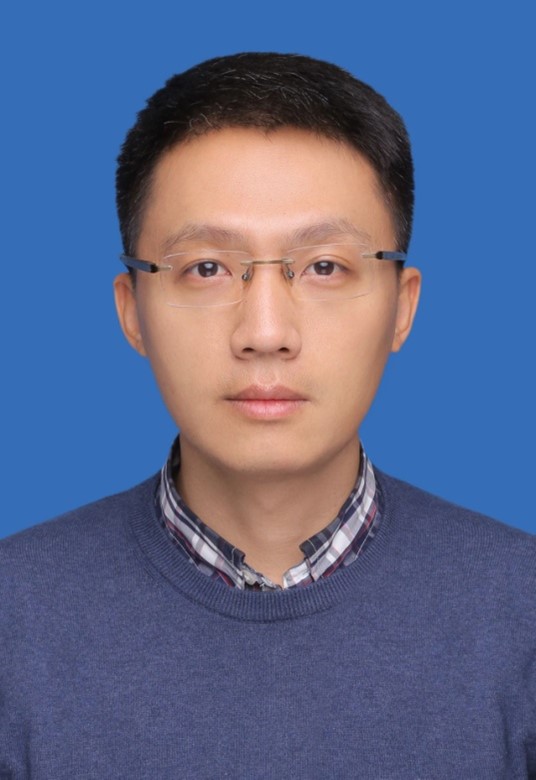}}]{Lifu Shi}
received the Master degree from the Jilin University, China, in 2013. He is mainly researching in the field of data statistics and software science. In recent years, his research mainly focuses on big data analysis and data reconstruction in the field of magnetic medical detection and treatment such as magnetic particle imaging, homogeneous magnetic sensitivity immunoassay, etc.,and also leading the development of several software systems for clinical diagnosis.
\end{IEEEbiography}

\vfill

\end{document}